\documentclass[letterpaper,10 pt, conference]{ieeeconf}  % Comment this line out if you need a4paper
\usepackage{cite}
\usepackage{amsmath,amssymb,amsfonts}
\usepackage{algorithmic}
\usepackage{graphicx}
\usepackage{textcomp}
\usepackage{xcolor}
\usepackage{flushend}

\IEEEoverridecommandlockouts                              % This command is only needed if 
\title{\LARGE \bf
Varying Joint Patterns and Compensatory Strategies Can Lead to the Same Functional Gait Outcomes: A Case Study
}

\author{Tomislav Bacek$^{1*}$, Mingrui Sun$^{1}$, Hengchang Liu$^{1}$, Zhongxiang Chen$^{2}$, Dana Kuli\'c$^{2}$, \\Denny Oetomo$^{1}$ and Ying Tan$^{1}$% <-this % stops a space
\thanks{This work was funded by the Australian Research Council, Project scheme (DP190100916).}% <-this % stops a space
\thanks{*The corresponding author: {\tt\small tbacek@unimelb.edu.au}}
\thanks{$^{1}$T. Bacek, M. Sun, H. Liu, D. Oetomo and Y. Tan are with the Faculty of Engineering and Information Technology, The University of Melbourne, Parkville 3010, VIC, Australia}%
\thanks{$^{2}$Z. Chen and D. Kulic are with the Faculty of Engineering, Monash University, Clayton 3800, VIC, Australia. }%
}

\begin{document}

\maketitle
\thispagestyle{empty}
\pagestyle{empty}
%%%%%%%%%%%%%%%%%%%%%%%%%%%%%%%%%%%%%%%%%%%%%%%%%%%%%%%%%%%%%%%%%%%%%%%%%%%%%%%%
\begin{abstract}
% two points to get across:
% (1) INDIVIDUALISATION - different compensatory strategies
% (2) TASK vs JOINT SPACE - quality vs functional gait params (redundancy)
% - - - - - - 
% OBJECTIVE
% (1) individualisation
This paper analyses joint-space walking mechanisms and 
% (2) redundancy
redundancies in delivering functional gait outcomes. 
% - - - - - - 
% METHODS
Multiple biomechanical measures are analysed for two healthy male adults who participated in a multi-factorial study and walked during three sessions. 
% - - - - - - 
% RESULTS
% (1) Individualisation
Both participants employed varying intra- and inter-personal compensatory strategies (e.g., vaulting, hip hiking) across walking conditions
% (2) Task vs joint space
and exhibited notable gait pattern alterations while keeping task-space (functional) gait parameters invariant. They also preferred various levels of asymmetric step length but kept their symmetric step time consistent and cadence-invariant during free walking.
% - - - - - - 
% CONCLUSION and SIGNIFICANCE
The results demonstrate the importance of an individualised approach and the need for a paradigm shift from functional (task-space) to joint-space gait analysis in attending to (a)typical gaits and delivering human-centred human-robot interaction.
\end{abstract}
%%%%%%%%%%%%%%%%%%%%%%%%%%%%%%%%%%%%%%%%%%%%%%%%%%%%%%%%%%%%%%%%%%%%%%%%%%%%%%%%
\section{INTRODUCTION}
Human walking has been characterised by well-established patterns \cite{Winter2009} believed to be strongly related to the gait economy \cite{Kuo2005}. However, much remains unknown about seemingly hierarchical \cite{Ivanenko2004a} walking mechanisms on an individual level due to the large inter/intra-person variability \cite{Winter2009,Fukuchi2018}. Understanding individual gait variations is crucial for enabling effective physical human-robot interaction (pHRI). This paper presents preliminary results of an extensive human study designed to address this major gap. 

Healthy adults have been reported to walk at \textit{their optimal} spatio-temporal gait symmetries \cite{Kodesh2012} and step lengths \cite{Donelan2001} and can continuously optimise their gait \cite{Selinger2015}. This has been exploited in the \textit{assistance-facilitated adaptation} (AFA) approach \cite{Zhang2017,Jackson2019,Ding2018}. In these works, a measured functional human performance (i.e., metabolic efficiency) was used to tune robotic input to improve that same performance within pHRI. Despite positive outcomes, the AFA approach is limited to non-clinical environments (e.g., power augmentation and assistance \cite{Young2016}) as it does not consider gait quality (i.e., joint space) and its benefits vanish when pHRI ends \cite{Bacek2021}.

A different approach -- \textit{assistance-induced (motor) learning} (AIL), is needed in rehabilitation, where the goal is to evoke long-term motor changes that remain even after pHRI ends. To achieve this, both human and robot need to be optimised (over time), as proposed by the co-adaptation paradigm \cite{Gallina2015}. However, optimising walking (to induce motor learning) is particularly challenging in the patient population. Hemiparetic patients tend to experience gait symmetry as an energetic penalty \cite{Roemmich2019}, which forces them to maintain asymmetry \cite{Hsu2003} and walk at not the most efficient speeds \cite{Roemmich2019,Finley2017}. Furthermore, the complexities and uniqueness of compensatory strategies \cite{Ali2014} often lead to more effortfull and less stable walking \cite{Farris2015,Kao2014}. This often limits conventional physical therapy, and consequently, a robot-assisted one, to focus on functional (task space -- e.g., metabolic cost, gait speed) rather than quality (joint space -- e.g., gait trajectory) aspects of walking. As a result, robot-assisted gait therapy is yet to show clear benefits over traditional approaches \cite{Mehrholz2020}. 

Despite seemingly different mechanisms in healthy and patient gait, some research suggests separation of the effects of speed and compensatory walking \cite{Hsu2003, Finley2017}, similar to the effects of speed in healthy cohorts \cite{Fukuchi2019}. Other research \cite{Stoquart2012, Falisse2019} suggests that hierarchical organisation of gait parameters driving impaired gait can be studied outside the patient population provided relevant walking constraints are in place. If true, this would significantly simplify efforts to gain knowledge about the overarching gait adaptations and compensatory strategies, a prerequisite for the uptake of the AIL approach and pHRI fully centred around human users. 

This paper presents preliminary results of a multi-factorial human study (on gait signature and compensatory mechanisms), demonstrating the importance of an individualised approach and the need to expand from functional gait quality metrics. In full, the study will provide a wealth of (shared) experimental data (with healthy adults), with the intended use in guiding computational modelling predicting AIL outcomes in pHRI and informing the design of a follow-up study investigating compensatory strategies in the stroke population. 
%%%%%%%%%%%%%%%%%%%%%%%%%%%%%%%%%%%%%%%%%%%%%%%%%%%%%%%%%%%%%%%%%%%%%%%%%%%%%%%%
\section{MATERIALS AND METHODS}
\subsection{Participants}
Two healthy male adults with no known gait-interfering impairments were recruited for the study (P1: 84 kg, 1.8 m; P2: 85 kg, 1.78 m). Both P1 and P2 had prior experience walking on a dual-belt treadmill, but only P1 with the knee orthosis (Sec. II.D). Both participants signed the informed consent form, and the ethics committee of the University of Melbourne approved the study (2021-20623-13486-3).
% *******************************
\subsection{Experimental conditions}
The study builds on three multidimensional factors: (1) walking speed has slow ($v_1$=0.4 m/s), medium ($v_2$=0.8 m/s), and normal ($v_3$=1.25 m/s) levels; (2) impairment is binary: free (w/o constraints, $c_1$) and impaired (w/ constraints, $c_2$) walking; and step frequency changes between preferred ($f_3$) and two lower ($f_1$=0.9$f_3$, $f_2$=0.95$f_3$) and higher ($f_4$=1.1$f_3$, $f_5$=1.2$f_3$) frequencies. Each participant goes through all 30 (2x3x5) factor combinations. Asymmetric step frequency allows consistency across walking speeds since reducing $f_3$ by 20\% at $v_1$ leads to highly imbalanced walking.

\begin{figure}[thpb]
      \centering
      %\framebox{\parbox{0.97\linewidth}{}}
      \includegraphics[scale=0.7]{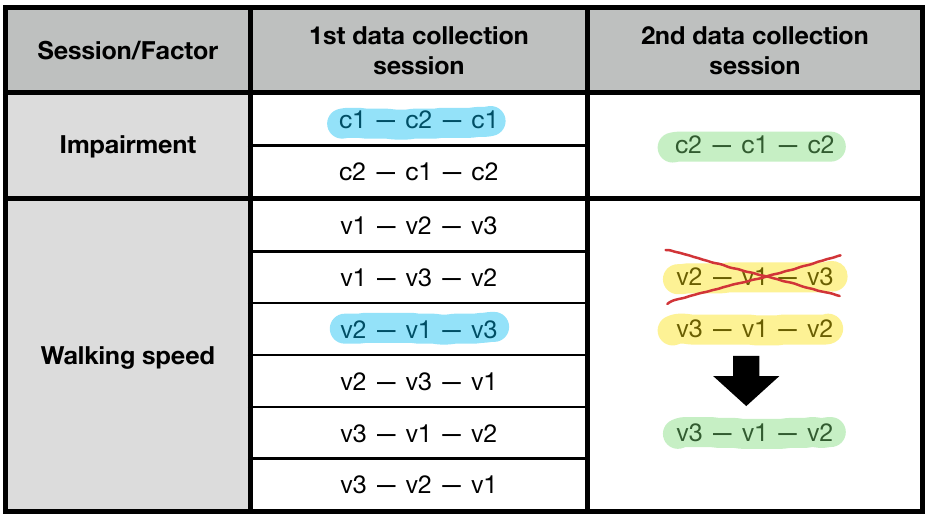}
      \caption{Example experimental conditions. The participant is randomly assigned to one impairment and one speed combination during the 1st day (blue), leaving one impairment combination for the 2nd day (green). The inverted impairment combination leaves two speed combinations available on the 2nd day (yellow). To avoid the order-of-speed bias (i.e., repeated speed) leaves only one speed combination (in green) to choose from.}
      \label{fig_ExpCond}
   \end{figure}

The six $v_i$-$c_j$ combinations are split across two data-collection sessions/days (Sec. II.C). To avoid familiarisation to speed or impairment, multiple occurrences of the speed (e.g., $v1$-$v2$-$v1$), or consecutive occurrences of the impairment condition (e.g., $c1$-$c2$-$c2$) within a session are rejected. When extended by the requirement to avoid order-of-speed bias, the choice of speed-impairment combinations across the two sessions is uniquely determined (see Fig. \ref{fig_ExpCond}).
% * * * * * * FIGURE * * * * * * 
% *******************************  
\subsection{Experimental protocol}
\begin{figure*}[ht!]
      \centering
     %\framebox{\parbox{6.7in}{}}
      \includegraphics[scale=0.5]{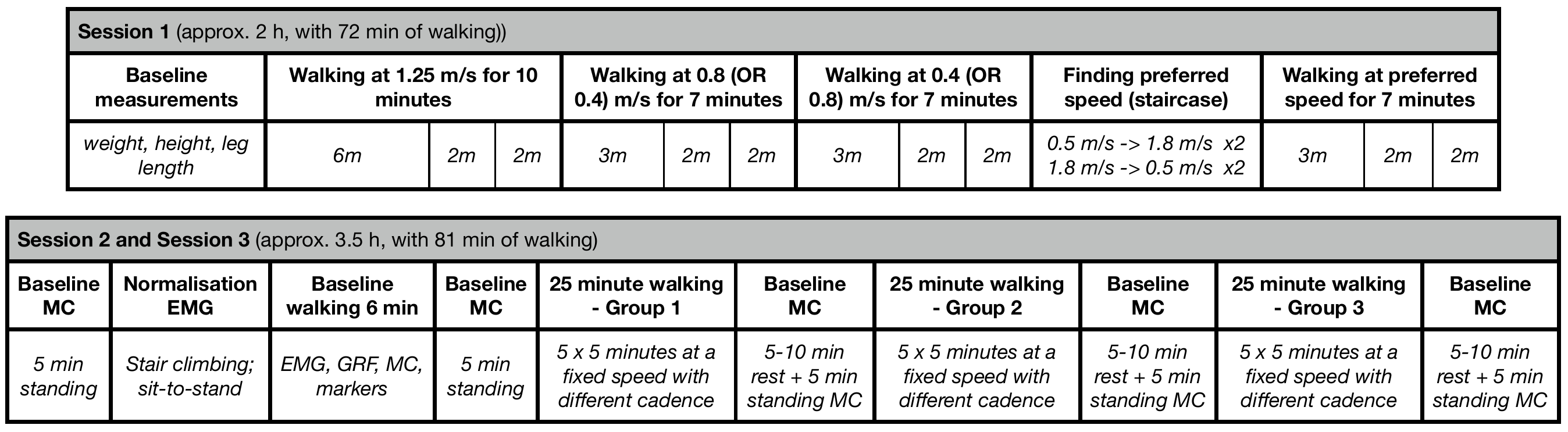}
      \caption{Experimental protocol. \textit{Top}: After taking their baseline measurements, participants walk at different speeds to determine their preferred cadences and comfortable speed. The self-selected cadence at either speed is an average of three values measured during the last 15 sec of a steady-state (6 or 3 min) and exploratory (2 min) blocks. During the latter, participants are guided by metronome away from their preferred frequency for 30 seconds and then left to walk as comfortable. No rest is given between the tests. The entire block is repeated twice, with and without the orthosis, totalling 72 minutes of walking (2x36). \textit{Bottom}: Participants start data collection sessions by walking at the preferred speed for six minutes, which serves as a warm-up and a baseline. Rest periods of 5-10 min are given and resting metabolic cost is measured in standing before and after each test. Participants are warned using auditory cues about the upcoming change in step frequency to ensure they are aware of the change. Across the two sessions, participants cover five step frequencies at all three walking speeds twice -- once with and once without the orthosis. Participants starting with two 25-minute bouts with the orthosis in Session 2 only have one such bout in Session 3, and vice versa. No participant walks more than 25 minutes in a single block.}
      \label{fig_study}
\end{figure*}

The study is organised into multiple sessions, bouts, and conditions. The participants walk on a treadmill on three days, each corresponding to one session. The first session (Ses1) is a preparatory day, while the second (Ses2) and third (Ses3) sessions are data collection days (Fig. \ref{fig_study}). 
% - - - - -
\subsubsection{Preparatory session}
This session serves to collect baseline data and for participants to get familiar with the study. They start by walking at 1.25 m/s and preferred cadence for six minutes \cite{Meyer2019}, followed by two 2-minute cadence-exploration periods to determine preferred step frequency \cite{Holt1991}. The same is repeated at 0.4 and 0.8 m/s.

The participants are then instructed to find their preferred walking speed using the staircase method \cite{Chung2010}. In short, the treadmill speed is twice gradually increased from 0.5 and decreased from 1.8 m/s, noting the participant's comfortable speed 4x and averaging it to get their preferred speed. The whole process of finding comfortable speed and step frequency at each speed is then repeated with the orthosis.
% - - - - -
\subsubsection{Data collection sessions}
Participants walk across three $v_i$-$c_j$ combinations lasting 25 minutes each during two data collection sessions. A metronome guides them to follow the desired step frequency that changes every five minutes in a randomised order. The first three minutes of each 5-min test allow participants to reach 95\% of their metabolic steady-state \cite{Selinger2015} and the last two are used for data averaging. Breaks of 5-10 minutes and a 5-min resting metabolic cost (MC) measurement separate the three 25-min bouts. Due to the paper's scope, MC analysis is not presented herein.

At the start of each session, participants are fitted with retroreflective markers for motion capture, wireless electromyogram (EMG) for muscle activity, and indirect calorimetry to measure MC. Their first walk at the preferred speed for six minutes serves as a warm-up \cite{Meyer2019} and a  baseline. This is followed by ten repeated motions of sitting down and two passes up and down a flight of 15 stairs to calculate the EMG normalisation factors \cite{Ghazwan2017}. 
% * * * * * * FIGURE * * * * * *
% - - - - -
\subsubsection{Walking speeds}
The 1.25 m/s is speed commonly used in tests with healthy participants \cite{Collins2015, Zhang2017, Malcolm2018} and is the preferred walking speed in young adults \cite{Winter2009, Fukuchi2019}. The other two speeds allow healthy-patients comparison in the future. The 0.8 m/s is a lower boundary of the community ambulator category \cite{Perry1995} and is typical of high-functioning hemiparetic patients \cite{Jonkers2009, Roemmich2019, Lewek2018}. The 0.4 m/s is an upper boundary of the household ambulator category \cite{Perry1995} and is typical of low-functioning hemiparetic patients \cite{Jonkers2009, Hornby2016, Mahtani2016}.
% *******************************  
\subsection{Unilateral orthosis}
\begin{figure}[ht!]
      \centering
      \includegraphics[scale=0.4]{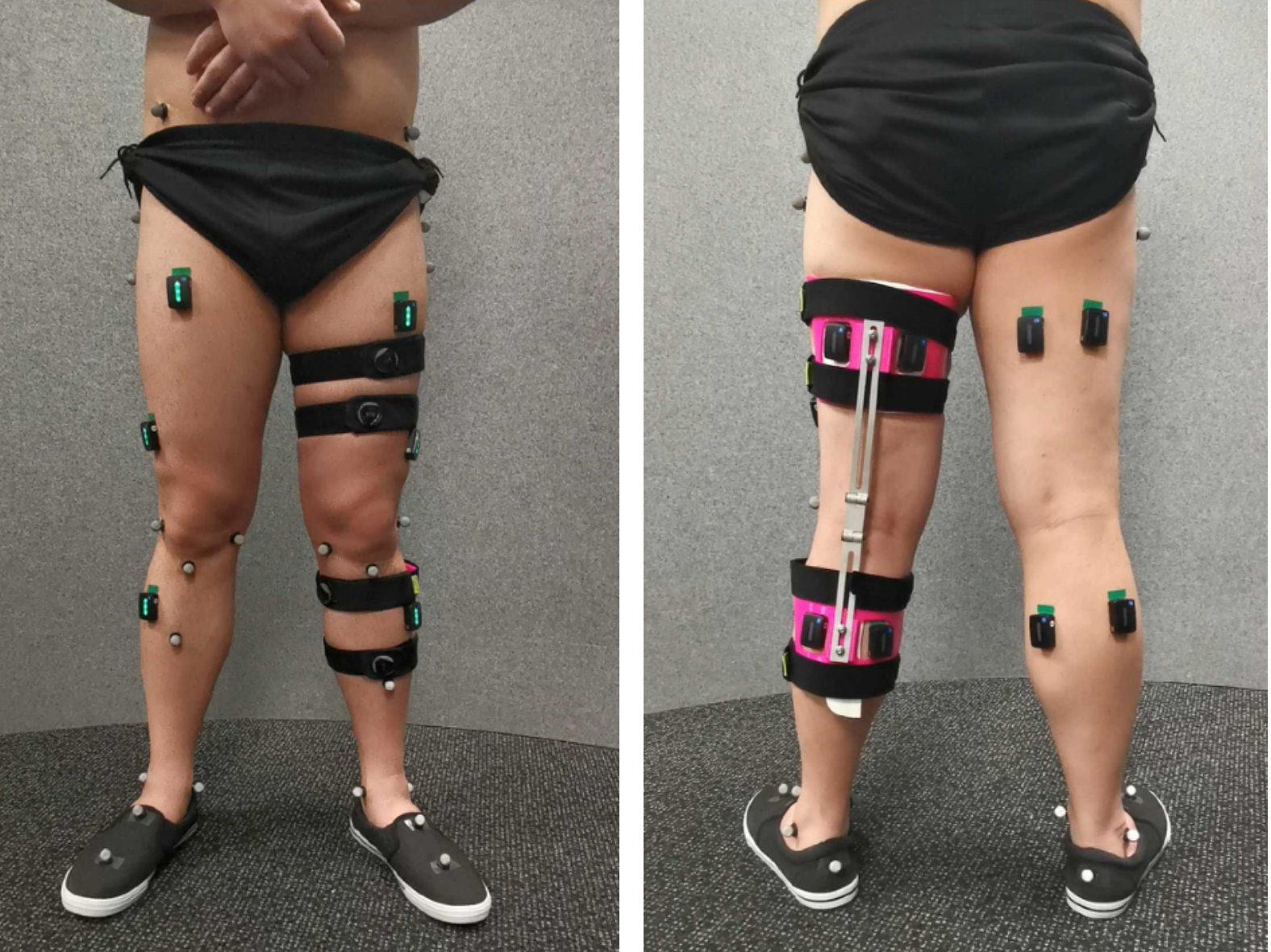}
      \caption{A unilateral knee orthosis. The orthosis is adjustable to different leg sizes and comes with cut-outs for the EMG sensors' placement. A 4 mm thick Softair White (Massons Healthcare, VIC, Australia) padding is placed on the cuffs' inner side to improve comfort and fitting. The cuffs are held in place by four BOA straps.}
      \label{fig_orthosis}
\end{figure}

A simple orthosis is used to lock the knee and elicit compensatory gait movements. The orthosis is worn on the left leg and consists of two 3D-printed cuffs and metal bars with a double-hinge joint (Fig. \ref{fig_orthosis}). The cuffs come in different sizes to account for variations in the participants' leg sizes, and their relative position to the knee joint can be changed with the grooved bars to provide a better fit and comfort. 
% * * * * * * FIGURE * * * * * * 
% *******************************
\subsection{Data collection}
The study is conducted using the Motek CAREN system located at the University of Melbourne. 
% - - - - -
\subsubsection{Kinematics and GRFs}
Gait data is collected using a dual-belt instrumented treadmill and 10 Vicon cameras (Vicon Motion Systems, Oxford, UK). Joint motion is collected at 100 Hz and ground reaction forces (GRFs) at 1 kHz. Each participant is fitted with a set of 26 markers (spherical, 14 mm diameter, B\&L Engineering, CA, USA) bilaterally placed in the pelvic and lower-limbs segments (custom marker template, Fig. \ref{fig_orthosis}).
% - - - - -
\subsubsection{Muscle activity}
The muscle activity of eight lower limb muscles per leg is measured using surface electromyography (sEMG) Delsys Trigno (Delsys Inc., Natick, MA, USA) system connected to Vicon for data sync purposes. The wireless Trigno Avanti sensors are placed in line with muscle fibers \cite{Konrad2006} over the Tibialis Anterior (TA), Gastrocnemius Lateralis (GL) and Medialis (GM), Vastus Lateralis (VL) and Rectus Femoris (RF), Biceps Femoris (BF) and Semitendinosus (ST), and Gluteus Maximus (GMax). The participant's skin is prepared following the SENIAM guidelines \cite{SeniamGuidelines}.
% *******************************
\subsection{Data processing}
All analyses are carried out using custom-written scripts in Matlab 2021a (Mathworks, Massachusetts, USA).
% - - - - -
\subsubsection{Joint motion and GRFs}
The kinematics data are filtered using a 4th order zero-lag Butterworth filter with 6 Hz cut-off frequency \cite{Winter2009}. Lower limb joint angles are calculated from filtered 3D marker trajectories as per \cite{Robertson2013} and following ISB guidelines \cite{Wu2002}. Data is segmented using GRFs and time-normalised to 0-100\% using linear interpolation (heel strike to subsequent heel strike). The static calibration trial performed while standing at the start of each session is used as a reference for neutral joint angles \cite{Robertson2013}.
% - - - - -
\subsubsection{Muscle activity}
Linear envelope detection in the raw EMG data is performed using a bandpass filter (10-500 Hz) followed by a full-wave rectification and smoothening with a 200 ms moving average window. Muscle activity collected during stair climbing, sit-to-stand, and baseline walking at the start of each session are pooled and used in finding normalisation factors (per muscle per person) to avoid major drawbacks of the isometric contraction approaches \cite{Ghazwan2017}.
% *******************************
\subsection{Parameter definitions}
\subsubsection{Step length and time symmetry}
Step length/time symmetry ($\Phi_T^s$,$\Phi_L^s$) is the ratio of the left leg to stride step length/time. Step length $\Phi_L$ is the fore-aft distance between the leading and trailing leg’s heel marker at the time of the leading leg’s heel strike. Step time $\Phi_T$ is the elapsed time between the same two events. A value of 50\% marks equal $\Phi_T$ or $\Phi_L$, while $>$50\% a longer left $\Phi_T$ or $\Phi_L$.
% - - - - -
\subsubsection{Hip hiking}
Hip hiking is the difference between the vertical position of the anterior superior iliac spine (SIS) marker during swing and standing. When the difference is positive, a person is considered to walk with hip hiking.
% - - - - -
\subsubsection{Pelvic obliquity}
Pelvic obliquity is a pelvic angle in the frontal plane. Pelvic obliquity on either leg is considered positive if the corresponding leg is lifted above the pelvic position measured in a standing trial.
%%%%%%%%%%%%%%%%%%%%%%%%%%%%%%%%%%%%%%%%%%%%%%%%%%%%%%%%%%%%%%%%%%%%%%%%%%%%%%%%
\section{RESULTS}
% *******************************
\subsubsection{Gait parameter symmetry}
\begin{figure*}[ht!]
      \centering
       \includegraphics[width=0.85\linewidth]{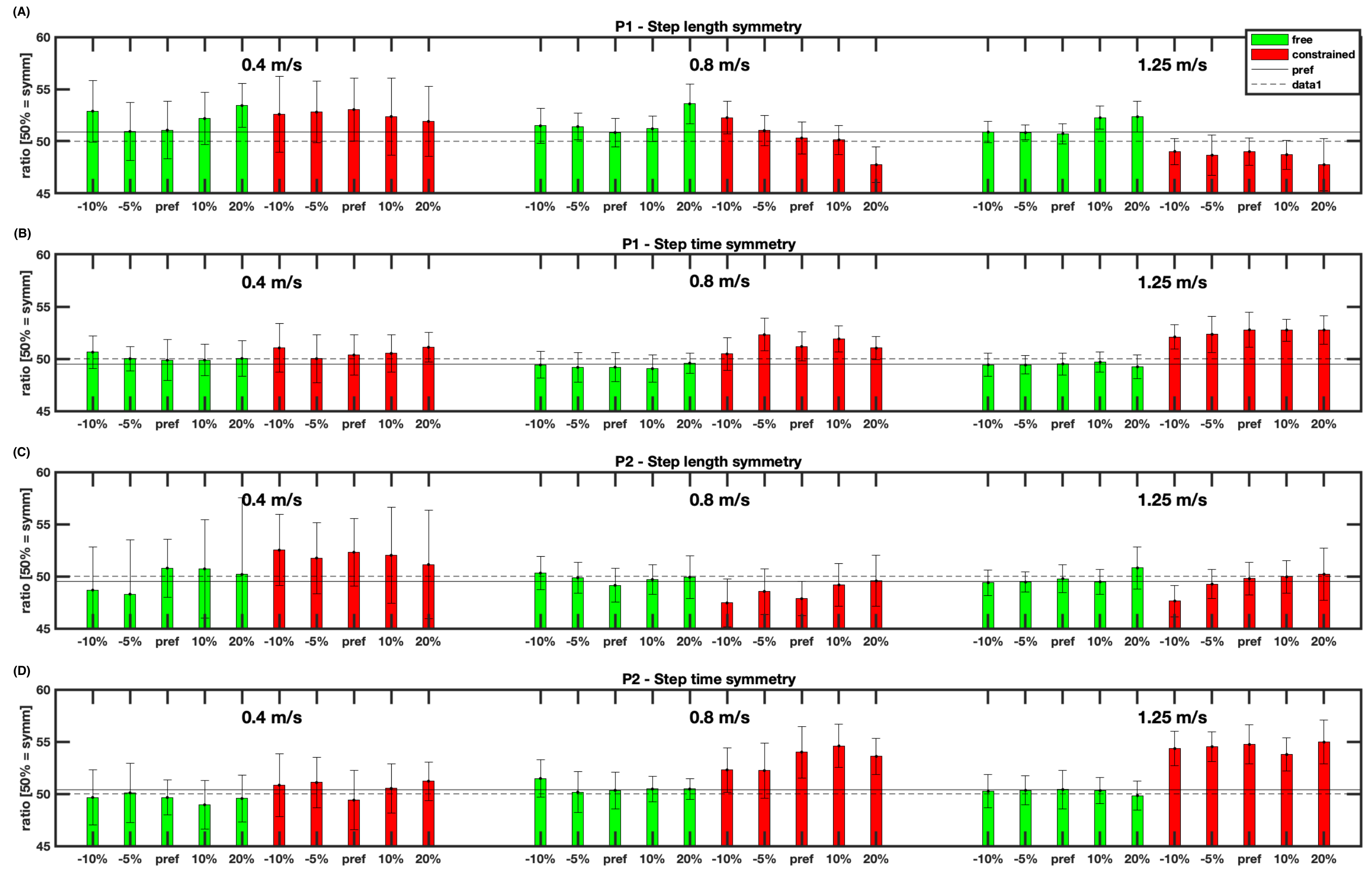}   
       \caption{Step length/time symmetry across all conditions for P1 and P2. Walking speeds are organised in columns and participants in rows (P1 in rows A and B, P2 in rows C and D). Green bars correspond to free and red bars to constrained walking. Each bar denotes a single step frequency. Bars represent the mean of the last minute of a 5-min test and error bars standard deviation. The solid horizontal line corresponds to a respective parameter value measured during preferred speed walking (P1: 1 m/s; P2: 1.05 m/s), averaged across two data collection sessions, while dashed horizontal line marks 50\% value.}
       \label{fig_symm} 
\end{figure*}

During free walking, P1 prefers asymmetric step length $\Phi_L$ and generally symmetric step time $\Phi_T$ (Fig. \ref{fig_symm}.A,B), while P2 generally walks symmetrically in both $\Phi_T$ and $\Phi_L$ (Fig. \ref{fig_symm}.C,D). Despite varying $\Phi_L^s$ (especially P1), the symmetry ratio at preferred step frequency $f_{pref}$ at all three speeds is the same as at $f_{pref}$ at comfortable walking speed. In general, the variability in both $\Phi_T^s$ and $\Phi_L^s$ decreases with an increase in speed.

Constraining the left knee generally reduced the constrained leg's $\Phi_L$ at 0.8 and 1.25 m/s (lower \% $\Phi_L^s$), and at multiple step frequencies also reversed the symmetry direction. A notable difference between the two participants is the effect of step frequency on $\Phi_L^s$: higher cadence decreases the constrained leg's $\Phi_L$ in P1 and increases in P2 at 0.8 and 1.25 m/s. On the other hand, both participants increased the left leg's $\Phi_T$ during constrained walking (higher \% $\Phi_T^s$), and more so with higher walking speeds.
% * * * * * * FIGURE * * * * * * 
% *******************************
\subsubsection{Compensatory strategies}
\begin{figure*}[ht!]
      \centering
      \includegraphics[width=0.8\linewidth]{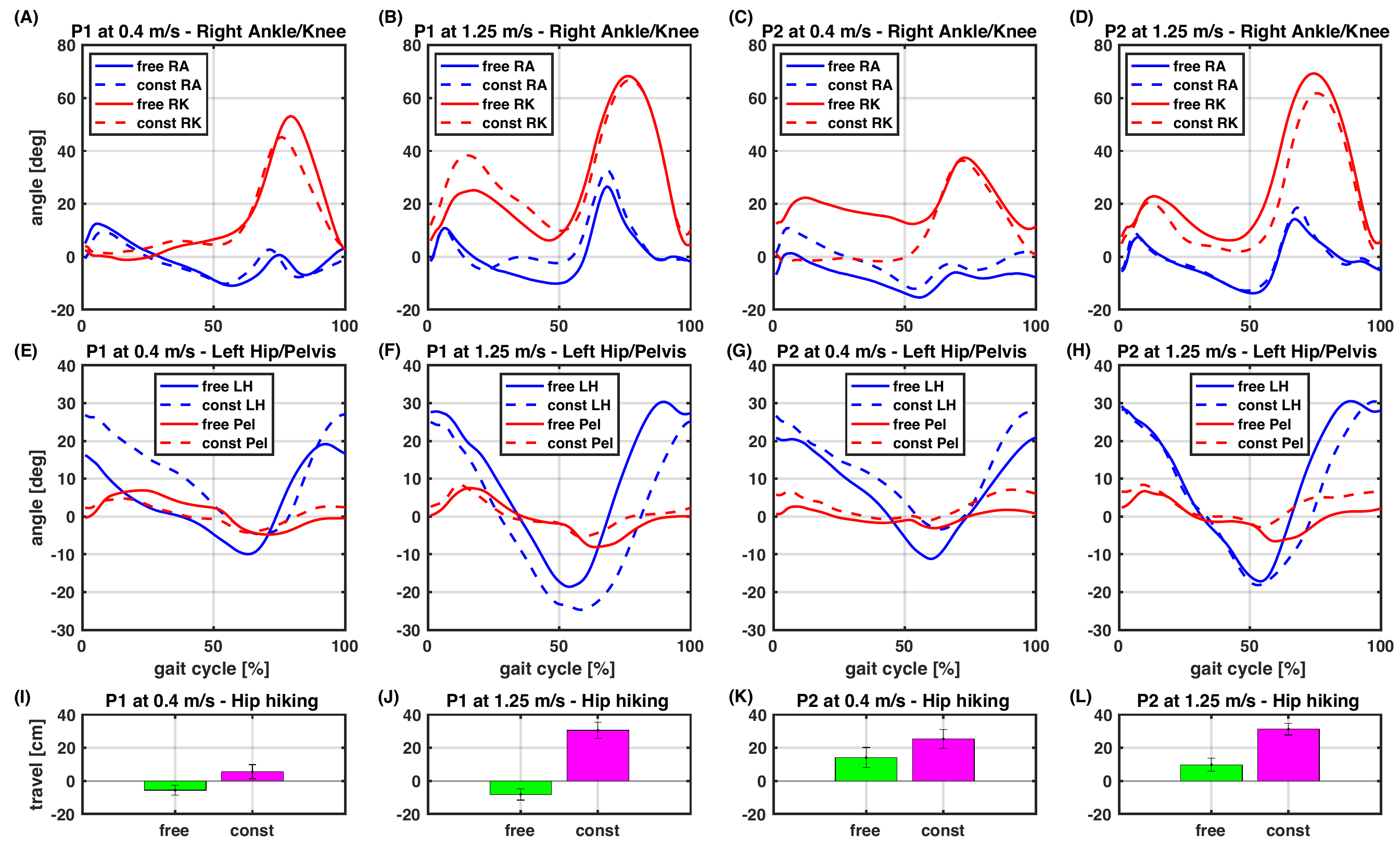}
      \caption{Compensatory strategies during walking at 0.4 and 1.25 m/s and preferred cadence. All trajectories and bars are colour-coded across the impairment factor. The top row shows the right (unconstrained) leg's ankle and knee joint trajectories ($RA$ = right ankle, $RK$ = right knee), and middle row pelvis and the left (constrained) leg's hip joint trajectories ($LH$ = left hip). The bottom row shows the participants' hip hiking. All data is an average over the last minute of walking of the respective 5-min test (standard deviation omitted for clarity purposes). Error bars in the bottom row show one standard deviation.}
      \label{fig_compens}
\end{figure*}

Constrained walking caused participants to increase their hip hike (Fig. \ref{fig_compens}.I-L), albeit using different strategies. At both 0.4 and 1.25 m/s (shown here), P1 relied on his right leg's and pelvic compensatory actions to swing his left leg forward. At 0.4 m/s, P1 increased his right knee extension and decreased right ankle plantarflexion (40-60\% gait, Fig. \ref{fig_compens}.A) and decreased pelvic obliquity to account for the left hip's delayed forward progression (Fig. \ref{fig_compens}.E). At 1.25 m/s, P1 exhibited no right ankle plantarflexion in mid-to-late stance (30-55\% gait, Fig. \ref{fig_compens}.B) -- called vaulting, to allow his constrained leg to swing forward. 

On the other hand, P2 mainly relied on his right knee and pelvis to allow the constrained leg to swing forward. At 0.4 m/s (Fig. \ref{fig_compens}.C), P2 locked his right knee throughout the stance phase and increased his pelvic rotation (Fig. \ref{fig_compens}.G) to allow ground clearance. At 1.25 m/s, P2 relied on his pelvis (Fig. \ref{fig_compens}.H) and right knee extension during mid-to-late stance (Fig. \ref{fig_compens}.D) to compensate for the constrained leg, and by moving his constrained leg during swing faster (Fig. \ref{fig_compens}.H).
% * * * * * * FIGURE * * * * * * 
% *******************************
\subsubsection{Joint-space gait alterations}
\begin{figure}[thpb]
      \centering
      %\framebox{\parbox{3in}{We suggest that you use a text box to insert a graphic (which is ideally a 300 dpi TIFF or EPS file, with all fonts embedded) because, in an document, this method is somewhat more stable than directly inserting a picture.}}
      \includegraphics[scale=0.32]{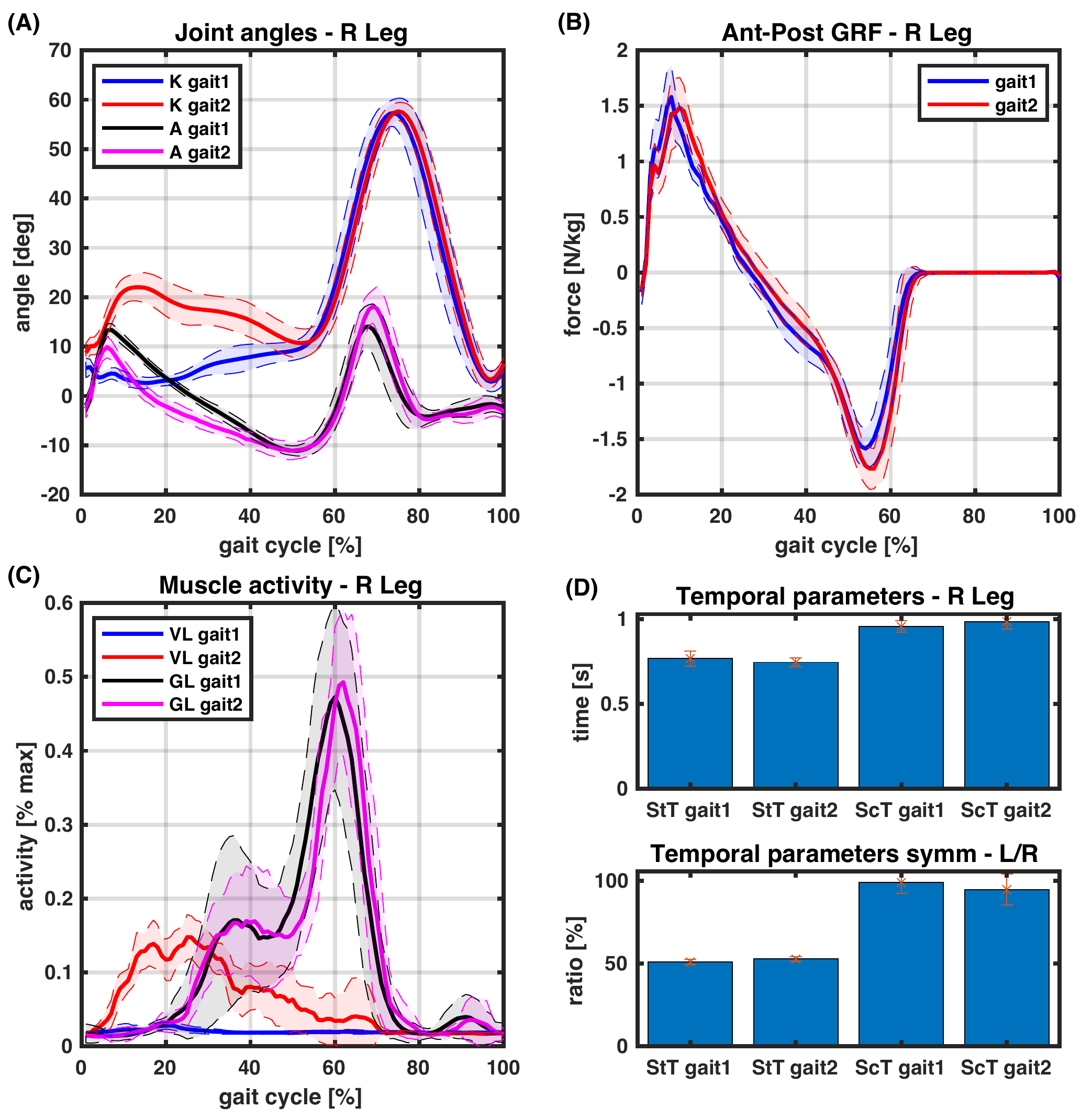}
      \caption{Joint-space patterns within a steady-state walking (P1 at 0.8 m/s, constrained and $f_{pref}$). Gait1 corresponds to cycles 23-79 and Gait2 cycles 80-93 within a 5-min test. All data are averaged over noted cycles and shown for the unconstrained leg, with the shaded area representing standard deviation. (\textbf{A}) Knee (K) and ankle (A) joint angles. (\textbf{B}) Anterior-posterior GRF corresponding to gait cycles in (A). (\textbf{C}) The measured activity of GastroLat (GL) and VasLat (VL) muscles. (\textbf{D}) Step time (StT) and stance time (ScT) of the unconstrained leg (\textit{Top}) and ratios to the left leg (\textit{Bottom}).}
      \label{fig_TwoGaits}
   \end{figure}

Both participants exhibited changes in joint trajectories during multiple 5-min bouts (fixed $v$, $c$, and $f$). With no external perturbations, their knee joint (left and/or right) sometimes switched multiple times between the locked and flexed knee in the stance phase (Fig. \ref{fig_TwoGaits}.A), also seen in the knee extensor EMG activity (Fig. \ref{fig_TwoGaits}.C). However, this affected neither the anterior-posterior GRF of the relevant leg (Fig. \ref{fig_TwoGaits}.B) nor the leg's $\Phi_T$ or $\Phi_T^s$ (Fig. \ref{fig_TwoGaits}.D).
% * * * * * * FIGURE * * * * * * 
%%%%%%%%%%%%%%%%%%%%%%%%%%%%%%%%%%%%%%%%%%%%%%%%%%%%%%%%%%%%%%%%%%%%%%%%%%%%%%%%
\section{DISCUSSION}
% - - - - - - -
% \subsection{Spatio-temporal gait symmetries}
In both participants and across all free walking conditions (Fig. \ref{fig_symm}), step time symmetry $\Phi_T^s$ remained largely invariant at  $\approx$50\%, similar to \cite{Kodesh2012}. Step length symmetry $\Phi_L^s$ -- different from \cite{Kodesh2012}, tended to settle over or under 50\% and was affected by speed and walking $f$. This is notable in P1, whose $\Phi_L$ asymmetry was minimal at $f_{pref}$ across all three speeds, matching his preferred walking speed's $\Phi_L^s$ (black line, Fig. \ref{fig_symm}). P2 showed the same tendency at 1.25 and 0.8 m/s. A higher consistency in $\Phi_T^s$ across conditions suggests that $\Phi_T^s$ might not play as dominant a role in shaping the energetic landscape of healthy persons' walking as suggested in \cite{Stenum2020}.

Constraining the left knee had both participants increase their $\Phi_L^s$ ratio at 0.4 m/s compared to free walking (i.e., longer constrained leg steps), with P1 keeping and P2 reversing his preferred asymmetry direction. This direction, opposite to what patients at comparable speeds prefer \cite{Finley2017}, is again reversed for both participants at 0.8 and 1.25 m/s. For P1, this also meant reversing his preferred $\Phi_L^s$ direction. Interestingly, increasing $f$ at the two speeds had the opposite effects on the $\Phi_L^s$ ratio in P1 (decreasing) and P2 (increasing). On the other hand, constraining the knee led to an increase in $\Phi_T^s$ at all speeds, and more so the higher the speed. At the same time, $\Phi_T^s$ shows more consistency across $f$ at higher walking speeds, which aligns with the theoretical importance timing of gait events has on walking energetics \cite{Kuo2005} and experimentally-validated time-invariance of key gait events during human-robot interaction \cite{Shirota2017, Bacek2021}. 

% - - - - - - -
% \subsection{Compensatory mechanisms}
The identified trends in spatio-temporal gait parameters, defined in task space, do not uniquely translate into the joint space due to redundancies in the human musculoskeletal system. For example, walking free at 0.4 m/s and preferred $f$ had P1 and P2 use different kinematics despite similar $\Phi_T^s$ and $\Phi_T^s$. P1 walked with no hip hiking, locked knee, and high dorsiflexion (Fig. \ref{fig_compens}.A,E,I), a complete opposite to P2 (Fig. \ref{fig_compens}.C,G,K). On the other hand, their joint trajectories look similar at 1.25 m/s. Interestingly, the timing of the key gait events (e.g., ankle push-off, peak knee flexion in swing) remained remarkably consistent, in line with theoretical considerations \cite{Kuo2005} and model predictions \cite{Koopman2014, Moissenet2019}. 

Constraining the knee increased hip hiking in both P1 and P2, albeit to a different extent (Fig. \ref{fig_compens}.I-L). Analogous to free walking, similar $\Phi_T^s$ and $\Phi_L^s$ at $f_{pref}$ translated into different compensatory strategies. At 0.4 m/s, P1 used corrective actions at the ankle and knee joint level to compensate for delayed forward progression of the constrained leg. P2, on the other hand, relied mostly on the increased pelvic tilt and locked knee joint. This strategy of relying on the free (i.e., unimpaired) leg when $\Phi_L^s>50\%$ is common in the patient population \cite{Allen2011}, which suggests that compensatory mechanisms may not only depend on the impairment but also the person's unique, pre-morbidity gait signature.

As Fig. \ref{fig_TwoGaits} shows, compensatory mechanisms are not the only way humans can adjust their gait. On an intra-person level, individuals can seemingly effortlessly change their gait in the joint space (joint trajectories) while not affecting the task space performance (GRF, $\Phi_T^s$). How these adjustments reflect in, e.g., gait energetics, remains unclear, as does why participants kept changing their joint trajectories. Invariance in GRFs and temporal gait parameters in gait trajectories points towards the preservation of the inverted pendulum dynamics \cite{Geyer2006}, a mechanism found in a neurologically-impaired population \cite{Stoquart2012} as well. However, the results presented in this paper demonstrate that human gait is much more complex than simple mathematical presentations and that it can only be understood by studying it on an individual level and in both task and joint space.
%%%%%%%%%%%%%%%%%%%%%%%%%%%%%%%%%%%%%%%%%%%%%%%%%%%%%%%%%%%%%%%%%%%%%%%%%%%%%%%%
\section{CONCLUSIONS}
This paper demonstrates through a case study that individuals do not necessarily optimise the same gait parameters and can utilise different compensatory strategies even when their spatio-temporal parameters are similar. The results also show that humans can seemingly effortlessly tap into the redundant musculoskeletal mechanisms, providing further evidence for the importance of studying both functional (task space) and quality (joint space) aspects of walking. 

%\addtolength{\textheight}{-1cm}   % This command serves to balance the column lengths
                                  % on the last page of the document manually. It shortens
                                  % the textheight of the last page by a suitable amount.
                                  % This command does not take effect until the next page
                                  % so it should come on the page before the last. Make
                                  % sure that you do not shorten the textheight too much.
%%%%%%%%%%%%%%%%%%%%%%%%%%%%%%%%%%%%%%%%%%%%%%%%%%%%%%%%%%%%%%%%%%%%%%%%%%%%%%%%
\section*{Appendix}
A video (.mp4) is added to this paper, showing P1 and P2's constrained walking at 0.4 and 1.25 m/s and preferred $f$ and demonstrating their different compensatory strategies.
%%%%%%%%%%%%%%%%%%%%%%%%%%%%%%%%%%%%%%%%%%%%%%%%%%%%%%%%%%%%%%%%%%%%%%%%%%%%%%%%
\section*{ACKNOWLEDGMENT}
The authors would like to thank Prof. Gavin Williams of the Epworth Hospital and the University of Melbourne (UoM), Prof. Jennifer McGinley of the Physiotherapy Department with the UoM, and Dr. Liuhua Peng of the School of Mathematics and Statistics with the UoM for their expertise in designing the study.
%%%%%%%%%%%%%%%%%%%%%%%%%%%%%%%%%%%%%%%%%%%%%%%%%%%%%%%%%%%%%%%%%%%%%%%%%%%%%%%%
%\section*{REFERENCES}
%\bibliography{HRI_2022} 
\bibliographystyle{IEEEtran}

\end{document}